\makeatletter\def\graphicscache@inhibit{true}\makeatother
\documentclass[runningheads]{llncs}
\usepackage[utf8]{inputenc}
\usepackage{graphicx} %
\usepackage{tabularx}
\usepackage{threeparttable}
\usepackage{amsmath,amssymb}
\usepackage{bm}
\usepackage{xspace}
\usepackage[per-mode=symbol,binary-units=true,range-units=single,range-phrase=-,detect-weight=true,detect-family=true]{siunitx}
\usepackage{subcaption}
\usepackage{stackengine}
\usepackage{subfiles}
\usepackage[TS1,T1]{fontenc}
\usepackage{tikz}
\usepackage[urlcolor  = blue]{hyperref}

\newcommand{\reffig}[1]{Fig.~\ref{#1}}
\newcommand{\reftab}[1]{Tab.~\ref{#1}}
\newcommand{\refsec}[1]{Sec.~\ref{#1}}

\DeclareSIUnit\pixel{px}

\begin{document}
\title{6D Object Pose Estimation using\\Keypoints and Part Affinity Fields}
\titlerunning{6D Object Pose Estimation using Keypoints and Part Affinity Fields}
\author{Moritz Zappel\and
Simon Bultmann \and
Sven Behnke}
\authorrunning{M. Zappel, S. Bultmann, and S. Behnke}
\institute{Autonomous Intelligent Systems, Computer Science Institute VI\\
University of Bonn, Germany\\
\email{s6mozapp@uni-bonn.de, \{bultmann,behnke\}@ais.uni-bonn.de}\\
\url{https://www.ais.uni-bonn.de}}
\maketitle              %
\begin{tikzpicture}[remember picture,overlay]
\node[anchor=north west,align=left,font=\sffamily,yshift=-0.2cm] at (current page.north west) {%
  In: Proceedings of 24th RoboCup International Symposium, June 2021.
};
\end{tikzpicture}%
\begin{abstract}
The task of 6D object pose estimation from RGB images is an important requirement for autonomous service robots to be able to interact with the real world.
In this work, we present a two-step pipeline for estimating the 6\,DoF translation and orientation of known objects. Keypoints and Part Affinity Fields (PAFs) are predicted from the input image adopting the OpenPose CNN architecture from human pose estimation.
Object poses are then calculated from 2D-3D correspondences between detected and model keypoints via the PnP-RANSAC algorithm.
The proposed approach is evaluated on the YCB-Video dataset and achieves accuracy on par with recent methods from the literature.
Using PAFs to assemble detected keypoints into object instances proves advantageous over only using heatmaps. Models trained to predict keypoints of a single object class perform significantly better than models trained for several classes.

\end{abstract}
\section{Introduction}
\label{sec:Introduction}
Object pose estimation is essential for autonomous robots to be able to interact with their environment. It has numerous real-world applications, such as robotic manipulation and human-robot interaction for autonomous service robots which are the focus of RoboCup@Home~\cite{cosero_frontiers}.

The task addressed in this work consists of detecting known objects and estimating their 6\,DoF orientation and translation in 3D space from a single RGB image.
In recent years, two-stage approaches, which first detect keypoints and then solve a Perspective-n-Point (PnP) problem to infer the object pose~\cite{6DoF,PVNet,BB8}, have been shown to provide robust and accurate results. However, keypoint detection remains difficult for occluded or truncated objects.

\begin{figure}[t]
 \centering
 \includegraphics[width=\textwidth]{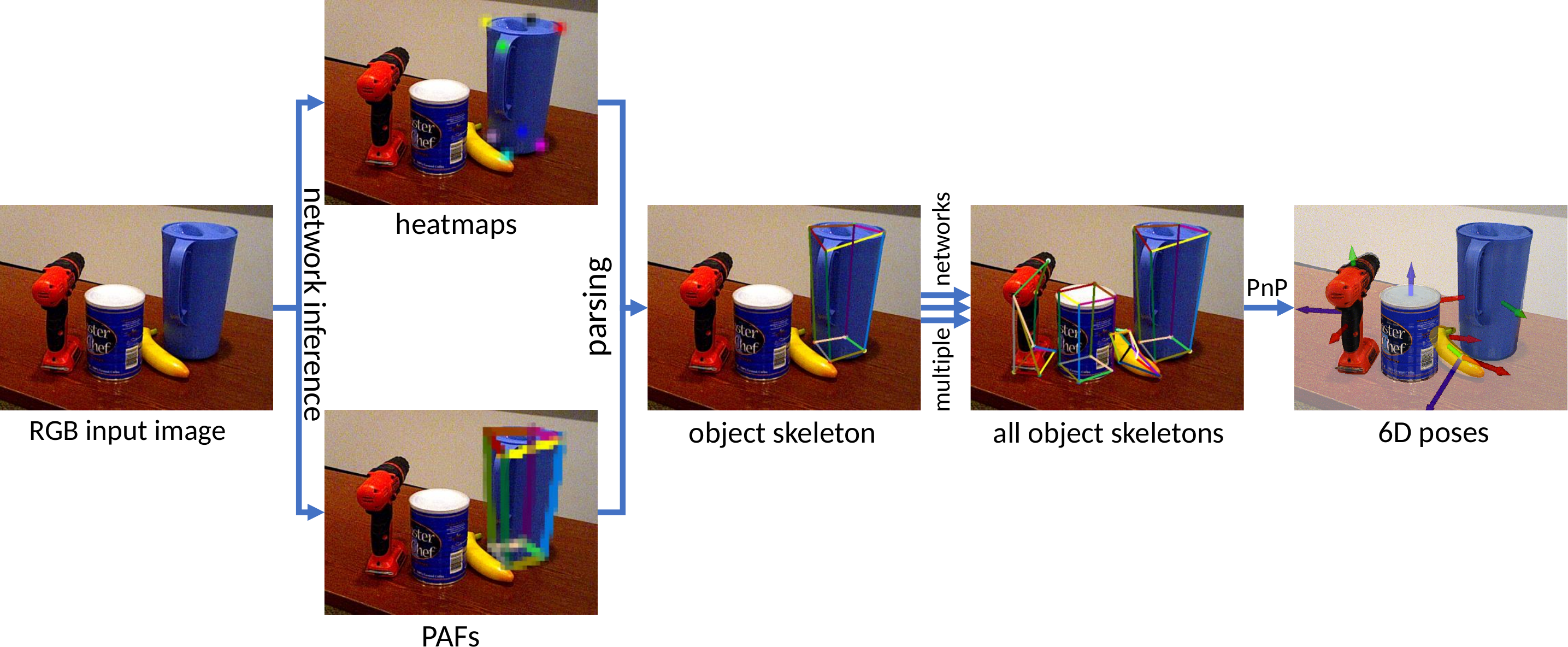}
 \vspace{-1.5em}
 \caption{Pose estimation approach: Heatmaps and PAFs are inferred for each object and 6D poses are calculated from detected keypoints via PnP.}
 \label{fig:workflow}
 \vspace{-1.5em}
\end{figure}

In this paper, we propose a two-stage pipeline for 6D object pose estimation in real-world scenes. We adopt the OpenPose architecture~\cite{OpenPose}, well known from person pose estimation, to detect keypoints and Part Affinity Fields (PAFs) of everyday objects. Keypoints are predicted as local maxima of a heatmap indicating the confidence of a part being present at the image location.
PAFs are vector fields connecting the keypoints of an object. They are used to assemble keypoints to object instances. As the OpenPose architecture is a bottom-up approach, keypoints are directly estimated from the input image. No prior object detection or segmentation is required, which is advantageous in terms of complexity and runtime~\cite{OpenPose}.
Correspondences between the predicted 2D keypoints from the image and the keypoints defined on the 3D object models are then used to calculate the 6\,DoF object poses via the PnP-RANSAC algorithm. An overview of the proposed method is given in \reffig{fig:workflow}.
Two different ways to define keypoints and PAFs on the object models are proposed in this work and the method is extensively evaluated on the challenging YCB-V dataset~\cite{PoseCNN}.

\section{Related Work}
\label{sec:Related_Work}
6D object pose estimation methods from the literature can roughly be divided into two classes.
Direct methods infer pose parameters directly from the image~\cite{PoseNet,PoseCNN}. PoseCNN~\cite{PoseCNN} defines a CNN architecture that segments objects in 2D images, predicts their depth, and regresses the 6D pose parameters. Capellen et al.~\cite{capellen_convposecnn_2020} extend this approach to a fully convolutional network for dense prediction of pose parameters not depending on prior object segmentation. The direct pose regression is, however, difficult---especially for the rotation parameters as the 3D rotation space is highly nonlinear.

In contrast, Keypoint-based approaches adopt a two-step pipeline: First, 2D keypoints are predicted for each object instance in the image and the 6\,DoF pose is computed in a second step from 2D-3D correspondences with a variant of the PnP-Algorithm~\cite{EPnP}. Approaches mainly differ in how the keypoints are defined on the object model and how they are inferred from the image. In BB8~\cite{BB8}, a segmentation mask is computed for each object and the keypoints are then inferred as the eight corners of the 3D object bounding box. The coordinates of the keypoints are directly regressed by the network. The corners of the 3D bounding box, however, often are not located on the object surface and are thus difficult to infer from local object image features.
Pavlakov et al.~\cite{6DoF} define keypoints on the object surface and infer them as maxima of pixel-wise heatmaps.
PVNet~\cite{PVNet} also defines keypoints on the object surface but infers them in a dense manner: Each pixel in the object segmentation mask predicts vectors that point to every keypoint. The keypoint locations are then computed through RANSAC-based voting, choosing locations where the predicted directions intersect. This permits to also represent keypoints that are occluded or outside of the image.

Some approaches apply additional refinement after the initial estimation of the pose to further improve performance. Cosypose~\cite{CosyPose}, e.g., implements an additional network that refines a given pose with the input image as additional input. When depth information is available, e.g., in the RGB-D variant of PoseCNN~\cite{PoseCNN}, ICP can also be used for pose refinement.

In this paper, we adopt a keypoint-based approach using keypoints on the object surface. The 6D object pose is computed via a combination of the PnP~\cite{EPnP} and RANSAC~\cite{Ransac} algorithms to increase the robustness of the estimation. Only RGB images are used as input and no further refinement steps are employed, keeping our pipeline simple and efficient.
The OpenPose architecture~\cite{OpenPose} is adopt\-ed for 2D keypoint estimation. OpenPose is a keypoint-based bottom-up approach for human pose estimation in images. Together with heatmaps of the keypoints, the CNN computes vector fields, called Part Affinity Fields (PAFs), connecting the keypoints of an object instance. This permits to directly predict keypoints on the input image without prior segmentation or detection required. Local maxima in the heatmaps are then assembled into instances via the PAFs. 
 
\section{Method}
\label{sec:Method}
In this section, we detail design choices and workflow of the proposed approach for 6\,DoF object pose estimation.

For evaluation, we employ the YCB-V data\-set~\cite{PoseCNN}. It comprises over 130k images at VGA resolution of 21 different object classes and is widely used for robot manipulation and object pose estimation tasks. For each object, a textured mesh is included as 3D model with the origin of the object coordinate frame defined at its center. All objects are household objects relevant for real-world robot experiments in domestic service scenarios (cf. \reffig{fig:heatmaps_YCB-V Objekte}). The images contain multiple objects in realistic settings with changing lighting conditions, significant image noise and cluttered backgrounds.
\vspace{-.7em}
\subsection{Selection of Keypoints and PAFs}
\begin{figure}[t]
\vspace{-.5em}
\centering
\captionsetup[subfigure]{labelformat=empty}
 \begin{subfigure}[b]{0.091\textwidth}
 \centering
  \includegraphics[width=\textwidth, height=3 cm,keepaspectratio]{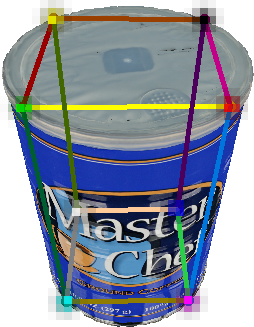}
  \caption{01}
  \label{fig:heatmaps_Objekt 01}
 \end{subfigure}
  \begin{subfigure}[b]{0.091\textwidth}
  \centering
  \includegraphics[width=\textwidth, height=3 cm,keepaspectratio]{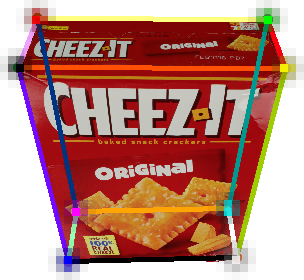}
  \caption{02}
  \label{fig:heatmaps_Objekt 02}
 \end{subfigure}
  \begin{subfigure}[b]{0.091\textwidth}
  \centering
  \includegraphics[width=\textwidth, height=3 cm,keepaspectratio]{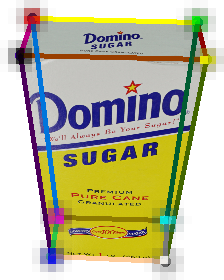}
  \caption{03}
\label{fig:heatmaps_Objekt 03}
 \end{subfigure}
  \begin{subfigure}[b]{0.091\textwidth}
  \centering
  \includegraphics[width=\textwidth, height=3 cm,keepaspectratio]{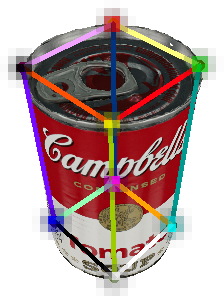}
  \caption{04}
  \label{fig:heatmaps_Objekt 04}
 \end{subfigure}
  \begin{subfigure}[b]{0.091\textwidth}
  \centering
  \includegraphics[width=\textwidth, height=3 cm,keepaspectratio]{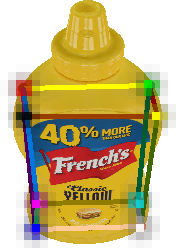}
  \caption{05}
  \label{fig:heatmaps_Objekt 05}
 \end{subfigure}
  \begin{subfigure}[b]{0.091\textwidth}
  \centering
  \includegraphics[width=\textwidth, height=3 cm,keepaspectratio]{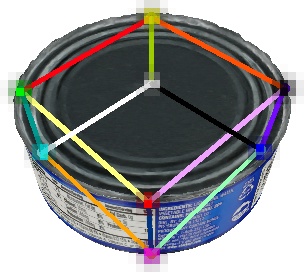}
  \caption{06}
  \label{fig:heatmaps_Objekt 06}
 \end{subfigure}
  \begin{subfigure}[b]{0.091\textwidth}
  \centering
  \includegraphics[width=\textwidth, height=3 cm,keepaspectratio]{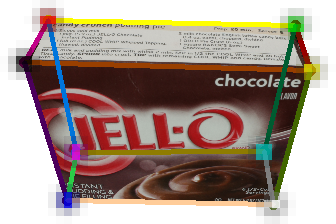}
  \caption{07}
  \label{fig:heatmaps_Objekt 07}
 \end{subfigure}
  \begin{subfigure}[b]{0.091\textwidth}
  \centering
  \includegraphics[width=\textwidth, height=3 cm,keepaspectratio]{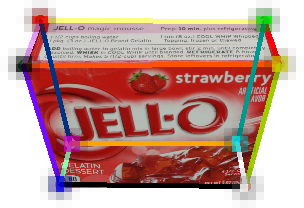}
  \caption{08}
  \label{fig:heatmaps_Objekt 08}
 \end{subfigure}
  \begin{subfigure}[b]{0.091\textwidth}
  \centering
  \includegraphics[width=\textwidth, height=3 cm,keepaspectratio]{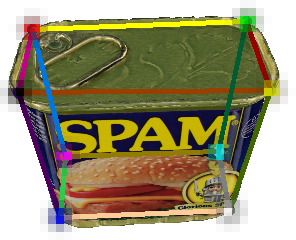}
  \caption{09}
  \label{fig:heatmaps_Objekt 09}
 \end{subfigure}
  \begin{subfigure}[b]{0.091\textwidth}
  \centering
  \includegraphics[width=\textwidth, height=1.5 cm,keepaspectratio]{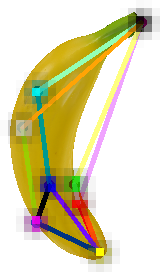}
  \caption{10}
  \label{fig:heatmaps_Objekt 10}
 \end{subfigure}

  \begin{subfigure}[b]{0.091\textwidth}
  \centering
  \includegraphics[width=\textwidth, height=3 cm,keepaspectratio]{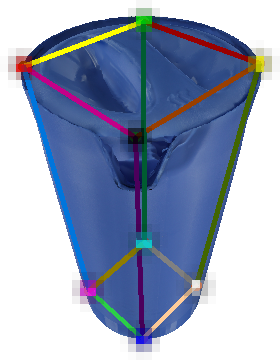}
  \caption{11}
  \label{fig:heatmaps_Objekt 11}  
 \end{subfigure}
  \begin{subfigure}[b]{0.091\textwidth}
  \centering
  \includegraphics[width=\textwidth, height=1.5 cm,keepaspectratio]{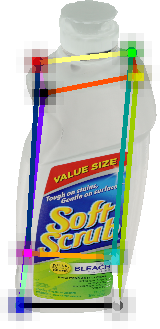}
  \caption{12}
  \label{fig:heatmaps_Objekt 12}
 \end{subfigure}
  \begin{subfigure}[b]{0.091\textwidth}
  \centering
  \includegraphics[width=\textwidth, height=3 cm,keepaspectratio]{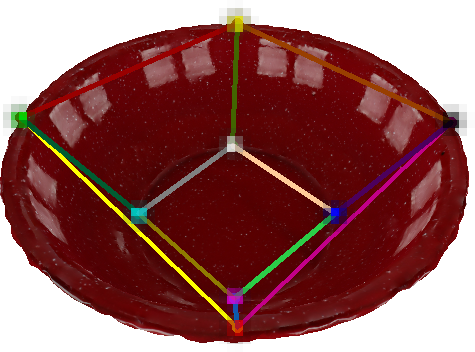}
  \caption{13}
  \label{fig:heatmaps_Objekt 13}  
 \end{subfigure}
  \begin{subfigure}[b]{0.091\textwidth}
  \centering
  \includegraphics[width=\textwidth, height=3 cm,keepaspectratio]{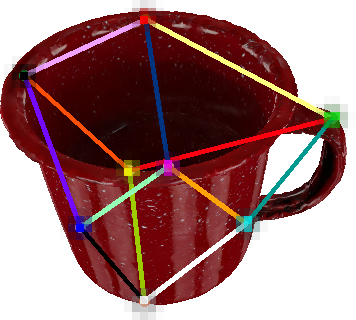}
  \caption{14}
  \label{fig:heatmaps_Objekt 14}
 \end{subfigure}
  \begin{subfigure}[b]{0.091\textwidth}
  \centering
  \includegraphics[width=\textwidth, height=3 cm,keepaspectratio]{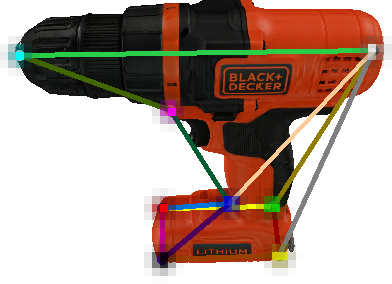}
  \caption{15}
  \label{fig:heatmaps_Objekt 15}
 \end{subfigure}
  \begin{subfigure}[b]{0.08\textwidth}
  \centering
  \includegraphics[width=\textwidth, height=2.2 cm,keepaspectratio]{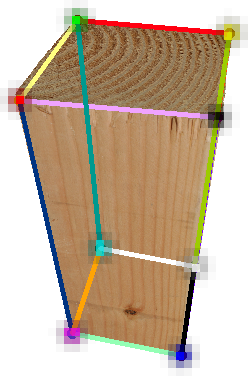}
  \caption{16}
  \label{fig:heatmaps_Objekt 16}  
 \end{subfigure}
  \begin{subfigure}[b]{0.08\textwidth}
  \centering
  \includegraphics[width=\textwidth, height=1.7 cm,keepaspectratio]{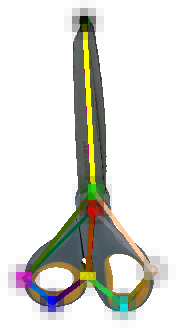}
  \caption{17}
  \label{fig:heatmaps_Objekt 17}  
 \end{subfigure}
  \begin{subfigure}[b]{0.065\textwidth}
  \centering
  \includegraphics[width=\textwidth, height=1.2 cm,keepaspectratio]{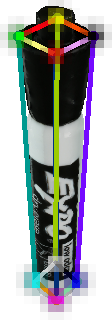}
  \caption{18}
  \label{fig:heatmaps_Objekt 18}  
 \end{subfigure}
  \begin{subfigure}[b]{0.08\textwidth}
  \centering
  \includegraphics[width=\textwidth, height=3 cm,keepaspectratio]{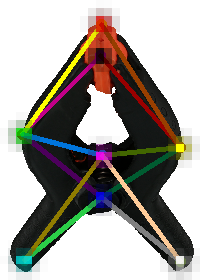}
  \caption{19}
  \label{fig:heatmaps_Objekt 19}  
 \end{subfigure}
  \begin{subfigure}[b]{0.08\textwidth}
  \centering
  \includegraphics[width=\textwidth, height=3 cm,keepaspectratio]{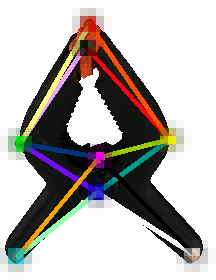}
  \caption{20}
  \label{fig:heatmaps_Objekt 20}  
 \end{subfigure}
  \begin{subfigure}[b]{0.08\textwidth}
  \centering
  \includegraphics[width=\textwidth, height=3 cm,keepaspectratio]{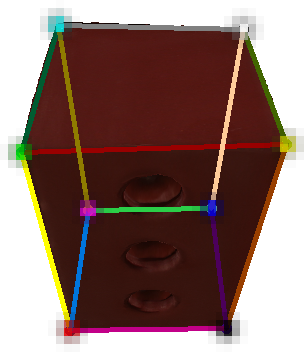}
  \caption{21}
  \label{fig:heatmaps_Objekt 21}  
 \end{subfigure}
 \caption{Objects of the YCB-V dataset with heatmaps of the manually defined keypoints and their interconnections.}
 \label{fig:heatmaps_YCB-V Objekte}
 \vspace{-1.5em}
\end{figure}

\begin{figure}[t]
\vspace{-.5em}
\centering
\captionsetup[subfigure]{labelformat=empty}
 \begin{subfigure}[b]{0.091\textwidth}
 \centering
  \includegraphics[width=\textwidth, height=3 cm,keepaspectratio]{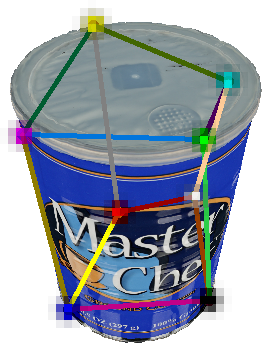}
  \caption{01}
  \label{fig:heatmaps_autoObjekt 01}
 \end{subfigure}
  \begin{subfigure}[b]{0.091\textwidth}
  \centering
  \includegraphics[width=\textwidth, height=3 cm,keepaspectratio]{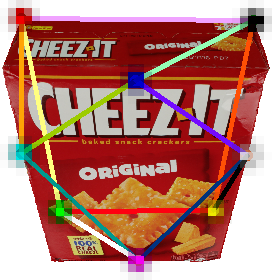}
  \caption{02}
  \label{fig:heatmaps_autoObjekt 02}
 \end{subfigure}
  \begin{subfigure}[b]{0.091\textwidth}
  \centering
  \includegraphics[width=\textwidth, height=3 cm,keepaspectratio]{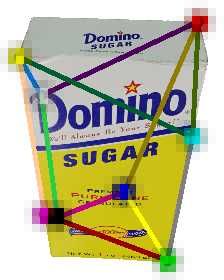}
  \caption{03}
\label{fig:heatmaps_autoObjekt 03}
 \end{subfigure}
  \begin{subfigure}[b]{0.091\textwidth}
  \centering
  \includegraphics[width=\textwidth, height=3 cm,keepaspectratio]{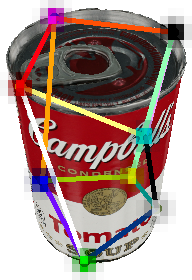}
  \caption{04}
  \label{fig:heatmaps_autoObjekt 04}
 \end{subfigure}
  \begin{subfigure}[b]{0.091\textwidth}
  \centering
  \includegraphics[width=\textwidth, height=3 cm,keepaspectratio]{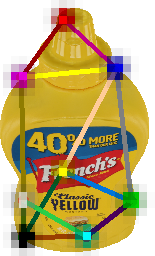}
  \caption{05}
  \label{fig:heatmaps_autoObjekt 05}
 \end{subfigure}
  \begin{subfigure}[b]{0.091\textwidth}
  \centering
  \includegraphics[width=\textwidth, height=3 cm,keepaspectratio]{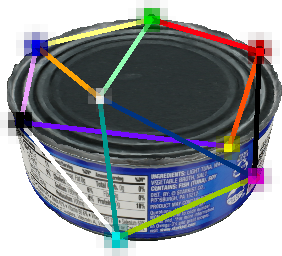}
  \caption{06}
  \label{fig:heatmaps_autoObjekt 06}
 \end{subfigure}
  \begin{subfigure}[b]{0.091\textwidth}
  \centering
  \includegraphics[width=\textwidth, height=3 cm,keepaspectratio]{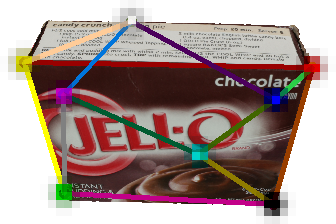}
  \caption{07}
  \label{fig:heatmaps_autoObjekt 07}
 \end{subfigure}
  \begin{subfigure}[b]{0.091\textwidth}
  \centering
  \includegraphics[width=\textwidth, height=3 cm,keepaspectratio]{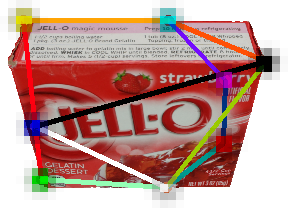}
  \caption{08}
  \label{fig:heatmaps_autoObjekt 08}
 \end{subfigure}
  \begin{subfigure}[b]{0.091\textwidth}
  \centering
  \includegraphics[width=\textwidth, height=3 cm,keepaspectratio]{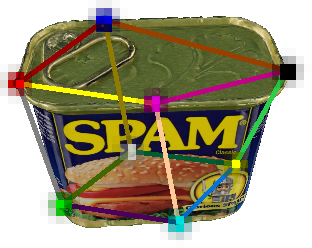}
  \caption{09}
  \label{fig:heatmaps_autoObjekt 09}
 \end{subfigure}
  \begin{subfigure}[b]{0.091\textwidth}
  \centering
  \includegraphics[width=\textwidth, height=1.5 cm,keepaspectratio]{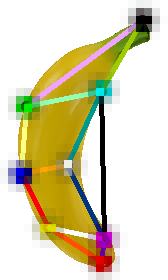}
  \caption{10}
  \label{fig:heatmaps_autoObjekt 10}
 \end{subfigure}

  \begin{subfigure}[b]{0.091\textwidth}
  \centering
  \includegraphics[width=\textwidth, height=3 cm,keepaspectratio]{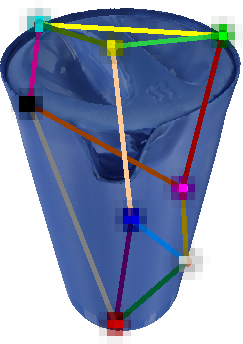}
  \caption{11}
  \label{fig:heatmaps_autoObjekt 11}  
 \end{subfigure}
  \begin{subfigure}[b]{0.091\textwidth}
  \centering
  \includegraphics[width=\textwidth, height=1.7 cm,keepaspectratio]{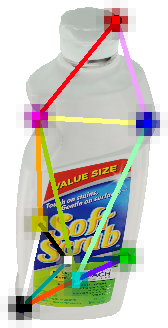}
  \caption{12}
  \label{fig:heatmaps_autoObjekt 12}
 \end{subfigure}
  \begin{subfigure}[b]{0.091\textwidth}
  \centering
  \includegraphics[width=\textwidth, height=3 cm,keepaspectratio]{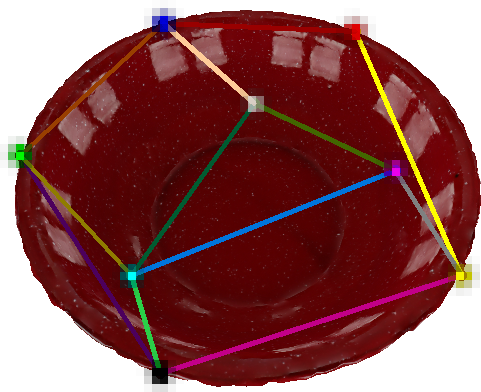}
  \caption{13}
  \label{fig:heatmaps_autoObjekt 13}  
 \end{subfigure}
  \begin{subfigure}[b]{0.091\textwidth}
  \centering
  \includegraphics[width=\textwidth, height=3 cm,keepaspectratio]{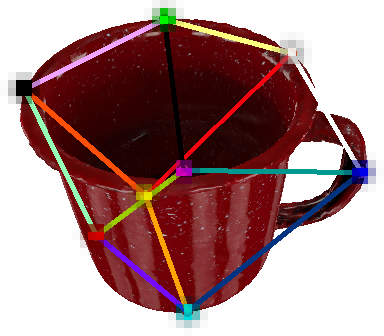}
  \caption{14}
  \label{fig:heatmaps_autoObjekt 14}
 \end{subfigure}
  \begin{subfigure}[b]{0.091\textwidth}
  \centering
  \includegraphics[width=\textwidth, height=3 cm,keepaspectratio]{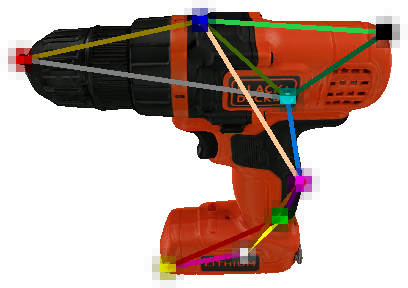}
  \caption{15}
  \label{fig:heatmaps_autoObjekt 15}
 \end{subfigure}
  \begin{subfigure}[b]{0.08\textwidth}
  \centering
  \includegraphics[width=\textwidth, height=2.2 cm,keepaspectratio]{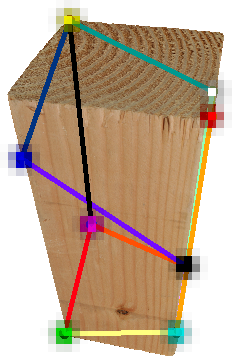}
  \caption{16}
  \label{fig:heatmaps_autoObjekt 16}  
 \end{subfigure}
  \begin{subfigure}[b]{0.08\textwidth}
  \centering
  \includegraphics[width=\textwidth, height=1.7 cm,keepaspectratio]{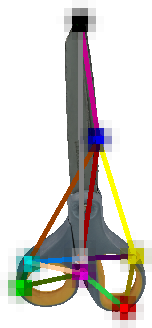}
  \caption{17}
  \label{fig:heatmaps_autoObjekt 17}  
 \end{subfigure}
  \begin{subfigure}[b]{0.065\textwidth}
  \centering
  \includegraphics[width=\textwidth, height=1.2 cm,keepaspectratio]{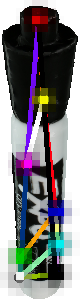}
  \caption{18}
  \label{fig:heatmaps_autoObjekt 18}  
 \end{subfigure}
  \begin{subfigure}[b]{0.08\textwidth}
  \centering
  \includegraphics[width=\textwidth, height=3 cm,keepaspectratio]{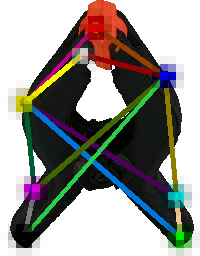}
  \caption{19}
  \label{fig:heatmaps_autoObjekt 19}  
 \end{subfigure}
  \begin{subfigure}[b]{0.08\textwidth}
  \centering
  \includegraphics[width=\textwidth, height=3 cm,keepaspectratio]{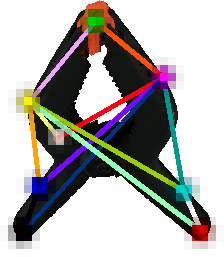}
  \caption{20}
  \label{fig:heatmaps_autoObjekt 20}  
 \end{subfigure}
  \begin{subfigure}[b]{0.08\textwidth}
  \centering
  \includegraphics[width=\textwidth, height=3 cm,keepaspectratio]{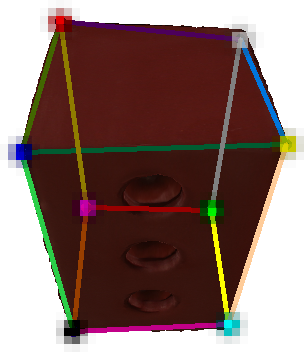}
  \caption{21}
  \label{fig:heatmaps_autoObjekt 21}  
 \end{subfigure}
 \caption{Objects of the YCB-V dataset with heatmaps of the automatically defined keypoints and their interconnections.}
 \label{fig:heatmaps_autoYCB-V Objekte}
 \vspace{-1.2em}
\end{figure}

The choice of keypoints and PAFs is an important design parameter of our method. They need to be well localized on the object geometry and texture, to facilitate their CNN-based detection, and should be spread out on the object surface such that a stable and well-defined solution of the PnP-problem can be found.
Keypoints and PAFs are defined based on the 3D object models in two different ways: They are chosen manually or automatically. Eight keypoints and twelve PAFs are defined per object class.
The manually defined keypoints are located on easy-to-find spots of the object geometry and texture and represent the object contour. If applicable, the keypoints are placed to form a cuboid. This set of keypoints is shown in \reffig{fig:heatmaps_YCB-V Objekte}.
The automatically defined set of keypoints is chosen with the farthest-point-algorithm, inspired by PVNet~\cite{PVNet}: Starting with the object center, points on the object surface which are farthest from the already chosen points are added to the keypoint set. Eight points on the object surface are retained---the center point is not part of the final keypoint set. The set of automatically chosen keypoints is shown in \reffig{fig:heatmaps_autoYCB-V Objekte}.
The PAFs for both sets of keypoints are defined by hand. The objective is to choose connections, which run along distinctive features and to form one upper and one lower polygon, which are connected with vertical PAFs. The PAFs run along the keypoint connections displayed in Figs.~\ref{fig:heatmaps_YCB-V Objekte} and~\ref{fig:heatmaps_autoYCB-V Objekte} and have a fixed width.
The automatically picked keypoints are less intuitively placed and harder to find than the manually picked ones. The reason for this is, that the automatically picked keypoints are often located on edges instead of corners and on surfaces instead of edges. Furthermore, the texture is ignored in the automatic selection although it is important to localize keypoints. The inferior performance of the automatically chosen keypoints is confirmed by the evaluation results (cf. \refsec{sec:Evaluation}). Therefore, the manually chosen keypoints are used for the main results of this paper.

The YCB-V dataset contains several symmetric objects: 13, 16, 19, 20, and 21. The bowl (Obj.~13) is rotationally symmetric while the other objects possess discrete symmetry transformations.
For each symmetric object, some poses are not distinguishable from each other. Hence, keypoints of the symmetric objects cannot be learned by the network if the symmetries are ignored. A simple elimination of symmetric poses during training is implemented in this work. All symmetry-equivalent poses are mapped to the same pose which ensures same relative position of keypoints on the image plane.

\subsection{Network}
We extend a public implementation of the OpenPose framework~\cite{tensorboy}. The network architecture is adjusted to the YCB-V dataset as well as the used keypoints and PAFs. For an input image of size $H \times W$ and $C$ object classes, the output shape is $\frac{H}{8} \times \frac{W}{8} \times ( C \cdot 8 + 1 )$ for the heatmaps and $\frac{H}{8} \times \frac{W}{8} \times ( C \cdot 12 \cdot 2 )$ for the PAFs. Heatmaps consist of 8 keypoint channels per object class and background, while PAFs consists of $x$ and $y$ channels for the 12 keypoint connections per class. Training labels are generated from the object model keypoints projected into the images of the video sequences using the annotated poses. A Gaussian blob is rendered at the keypoint position in the respective heatmap channel and the respective PAF channels represent the unit vector in the direction of the connection, within a fixed width along the connecting line, as in~\cite{OpenPose}.

To enable the network to estimate the poses for multiple object classes, the layer width of the intermediate stages needs
to be scaled accordingly to the output layers. A network for all 21 object classes would require more GPU memory than most graphic cards possess. Because of this, we train a separate model for each object class. We verify in the evaluation (\refsec{sec:Evaluation}), that the performance of these 1-object models is superior to models trained for multiple object classes.

\subsection{Workflow}
\reffig{fig:workflow} illustrates the general workflow of our pipeline for 6D object pose estimation. The input image is processed by the network and heatmaps and PAFs are estimated. The local maxima of each heatmap are candidates for the respective keypoint. These keypoint candidates are grouped into object instances using the PAFs, as in the OpenPose framework~\cite{OpenPose}. This step is repeated for every object class. The PnP and RANSAC algorithms are used to calculate the 6D poses of the found object instances with four or more valid keypoints\footnote{The PnP algorithm requires at least four correspondences for a unique solution.} using correspondences between detected 2D image keypoints and 3D model keypoints.
In datasets with only one instance of an object per image, the best guess will be kept. In multi-instance datasets, each estimated object pose will be assigned to the closest ground truth object pose of this class for evaluation.
 
\section{Evaluation}
\label{sec:Evaluation}
We evaluate our approach on the YCB-Video dataset~\cite{PoseCNN}, compare the results to other approaches from the literature, and perform ablation studies to understand the influences of different components of our method. All experiments run on a workstation PC with RTX 2080 GPU, i7-8700K CPU, and 32\,GB of RAM. Pose estimation takes \SI{50}{\milli\second} in average per object and image, thereof \SI{8}{\milli\second} for pre-processing, \SI{8}{\milli\second} for inference, and \SI{34}{\milli\second} for post-processing and PnP. The network requires \SI{1.24}{\giga\byte} of GPU memory.
\subsection{Metrics}
We employ two standard metrics for evaluation: average 3D distance of model points (ADD)~\cite{hinterstoisser2012} and 2D projection error~\cite{brachmann2016}. Both metrics employ the meshes of the object models to calculate the pose error. The ADD metric is defined as:
\begin{align}
\epsilon_{\text{ADD}} = \frac{1}{\vert P \vert} \sum_{\vec{p} \in P} \left.\Vert (\vec{R}\vec{p}+\vec{t})-(\tilde{\vec{R}}\vec{p}+\tilde{\vec{t}})\right.\Vert\,,
 \label{eq:add}
\end{align}
with $\tilde{\vec{R}}$ and $\tilde{\vec{t}}$ being the estimated rotation and translation, $\vec{R}$ and $\vec{t}$ defining the ground-truth pose and $P$ the set of vertices of the object model mesh.
For symmetric objects, the point-to-point correspondences can be ambiguous and the metric is adapted to compute the average distance using the closest point from the mesh~\cite{PoseCNN}:
\begin{align}
\epsilon_{\text{ADD-S}} = \frac{1}{\vert P \vert} \sum_{\vec{p}_1 \in P} \min_{\vec{p}_2 \in P} \left.\Vert (\vec{R}\vec{p}_1+\vec{t})-(\tilde{\vec{R}}\vec{p}_2+\tilde{\vec{t}})\right.\Vert\,.
\label{eq:adds}
\end{align}
The 2D projection metric computes the average 2D pixel distances between corresponding points projected onto the image plane of the evaluated view:
\begin{align}
\epsilon_{\text{2DProj}} = \frac{1}{\vert P \vert} \sum_{\vec{p} \in P} \left.\Vert \text{proj}(\vec{R}\vec{p}+\vec{t})-\text{proj}(\tilde{\vec{R}}\vec{p}+\tilde{\vec{t}})\right.\Vert\,.
 \label{eq:rep_er}
\end{align}
The evaluation scores are given in terms of the area under the accuracy-threshold curve (AuC). For this, the threshold for the respective distance metric is varied and the pose accuracy is computed for each threshold value. The maximum thresholds are set to \SI{10}{\centi\meter} for ADD(-S) and \SI{40}{\pixel} for the 2D projection metric.
\subsection{Results on the YCB-Video Dataset}
\begin{table}[t]
\centering
\caption{Area under accuracy Curve (AuC) for pose estimation of YCB-V objects. $^*$~denotes symmetric objects. Best results are marked bold.}
\setlength{\tabcolsep}{7pt}
\begin{tabular}{|r||r|r|r||r|r|}
\hline
Object & \multicolumn{3}{c||}{Ours} & \multicolumn{2}{c|}{PoseCNN~\cite{PoseCNN}} \\
\hline
&ADD&ADD-S&2D Proj.&ADD&ADD-S\\
\hline
 1&49.9&80.7&54.8& \textbf{ 50.9}& \textbf{ 84.0} \\ 
 \hline 
 2& \textbf{ 80.5}& \textbf{ 88.4}&84.3&51.7&76.9\\ 
 \hline 
 3& \textbf{ 85.5}& \textbf{ 92.4}&88.8&68.6&84.3\\ 
 \hline 
 4& \textbf{ 68.5}& \textbf{ 81.4}&84.8&66.0&80.9\\ 
 \hline 
 5& \textbf{ 87.0}& \textbf{ 93.3}&89.8&79.9&90.2\\ 
 \hline 
 6& \textbf{ 79.3}& \textbf{ 89.7}&81.7&70.4&87.9\\ 
 \hline 
 7& \textbf{ 81.8}& \textbf{ 89.5}&88.7&62.9&79.0\\ 
 \hline 
 8& \textbf{ 89.4}& \textbf{ 94.0}&92.9&75.2&87.1\\ 
 \hline 
 9&\textbf{59.6} &70.0&69.0& \textbf{ 59.6}& \textbf{ 78.5} \\ 
 \hline 
10&36.5&58.3&55.0& \textbf{ 72.3}& \textbf{85.9} \\ 
 \hline 
11& \textbf{78.1}& \textbf{86.9}&78.0&52.5&76.8\\ 
 \hline 
12& \textbf{56.7}&67.1&66.2&50.5& \textbf{71.9} \\ 
 \hline 
$^*$13&\textbf{12.2}&23.5&4.1&6.5&\textbf{69.7} \\ 
 \hline 
14&54.0&76.9&75.2& \textbf{57.7}& \textbf{78.0} \\ 
 \hline 
15& \textbf{82.8}& \textbf{91.0}&88.2&55.1&72.8\\ 
 \hline 
$^*$16&16.7&29.6&29.5&\textbf{31.8}& \textbf{65.8} \\ 
 \hline 
17& \textbf{46.0}& \textbf{64.1}&76.7&35.8&56.2\\ 
 \hline 
18&9.8&11.9&20.8& \textbf{58.0}& \textbf{71.4} \\ 
 \hline 
$^*$19&20.0&47.4&8.9& \textbf{25.0}& \textbf{49.9} \\ 
 \hline 
$^*$20&14.1&45.5&3.5& \textbf{15.8}& \textbf{47.0} \\ 
 \hline 
$^*$21&12.1&29.7&2.3& \textbf{40.4}& \textbf{87.8} \\ 
\hline
\hline
average& \textbf{59.0}& 72.7& 65.0&53.7& \textbf{75.9} \\
\hline
\end{tabular}
\label{tab:meine vs posecnn}
\vspace{-1.5em}
\end{table}
In \reftab{tab:meine vs posecnn}, we give detailed evaluation results of the AuC scores for all 21 objects of the YCB-V dataset and compare them to the results of PoseCNN~\cite{PoseCNN}. The proposed approach outperforms PoseCNN in terms of ADD for most of the non-symmetric objects and on average over all objects. The improvement is most significant for the box-shaped Objects 2, 3, 7, and 8 as well as for Objects 11, 15, and 17 which have a more complex shape (cf. \reffig{fig:heatmaps_YCB-V Objekte}).
PoseCNN, on the other hand, achieves better results for the symmetric objects in terms of ADD-S. The proposed approach provides less accurate results for these objects, where several keypoint configurations can result in visually equivalent poses. This makes the keypoint estimation harder to learn and cannot be fully compensated by the symmetry handling during training (cf. \refsec{sec:Method}). Also, keypoints are difficult to infer for objects with little prominent geometric features (e.g., edges or corners) such as Objects 10 and 18.
The 2D projection metric scores per object are not reported by the authors of PoseCNN.

In \reftab{tab:bop challenge vergleich}, we further compare the overall results of our method with the recent Benchmark for 6D Object Pose Estimation (BOP) challenge 2020~\cite{BOP,BOP2020}. The BOP challenge defines slightly different evaluation metrics: The MSSD and MSPD metrics are similar to the ADD and 2D projection metrics, but give the maximum value instead of the average error and deal with symmetries. The VSD metric is the percentage of pixels, which are visible in the estimated and ground truth pose and are close in the image space. The formal definitions are given in~\cite{BOP2020}.
We achieve the third-best result. In comparison to CosyPose~\cite{CosyPose}, which achieves the best result by a significant margin, we do not refine the initially estimated pose, which could further improve our result.

\begin{figure}[t]
 \centering
\begin{subfigure}[t]{0.49\textwidth}
  \includegraphics[width=0.49\textwidth]{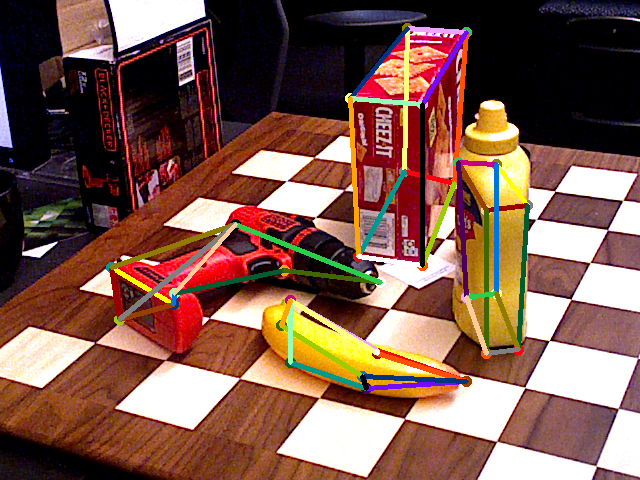}
  \includegraphics[width=0.49\textwidth]{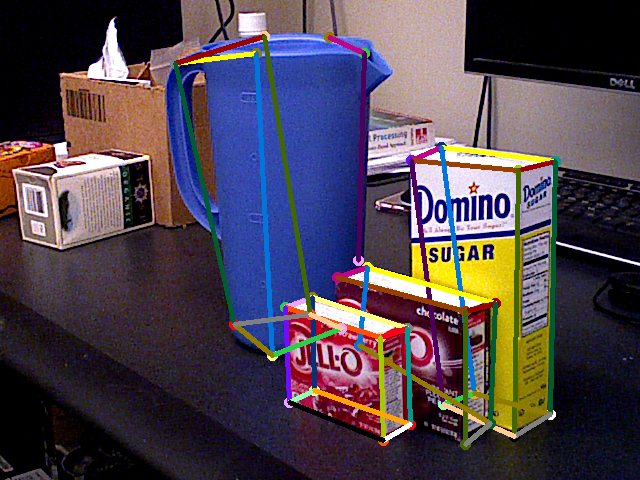}
  \caption{Object skeletons}
  \label{fig:best_imgs_vis}
 \end{subfigure}
\begin{subfigure}[t]{0.49\textwidth}
  \includegraphics[width=0.49\textwidth]{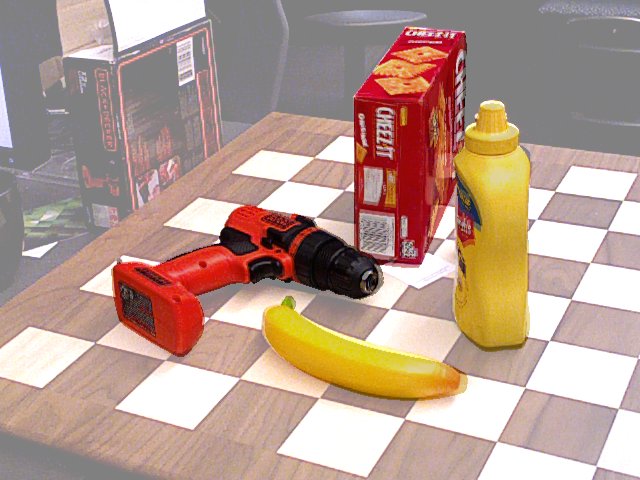}
  \includegraphics[width=0.49\textwidth]{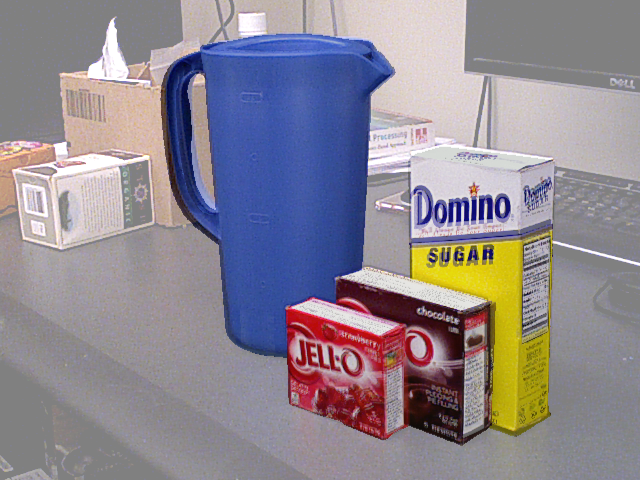}
  \caption{6D Poses}
  \label{fig:best_imgs_pnp}
 \end{subfigure}
 \vspace{-0.7em}
 \caption{Qualitative results on the YCB-V dataset: (a) keypoints and connections, (b) transformed object models overlay on input image.}
 \label{fig:best_imgs}
 \vspace{-1.7em}
\end{figure}

In \reffig{fig:best_imgs}, qualitative results on the YCB-V dataset are shown. 6D poses are estimated accurately despite occlusions and outlier keypoint detections.
\begin{table}[t]
\centering
\setlength{\tabcolsep}{4pt}
\caption{Results on YCB-V of the five best candidates using only RGB images of the BOP challenge 2020~\cite{BOP2020} in comparison with our work.}
\begin{tabular}{|l|c|c|c|c|c|}
\hline
Name & $AR$ & $AR_{VSD}$ & $AR_{MSSD}$ & $AR_{MSPD}$ & training data \\    \hline
CosyPose~\cite{CosyPose} &	0.821 &	0.772 &	0.842 &	0.850 & pbr+real \\    \hline
EPOS~\cite{EPOS} &	0.696 &	0.626 &	0.677 &	0.783 & pbr \\    \hline
\textbf{Ours} & 0.575 & 0.506 & 0.567 & 0.654 & pbr+real \\    \hline
CosyPose~\cite{CosyPose} & 0.574 & 0.516 & 0.554 & 0.653 & pbr \\ \hline
Leaping from 2D to 6D~\cite{leaping} & 0.543 & 0.443 & 0.499 & 0.687 & pbr+real \\    \hline
CDPNv2~\cite{CDPN} &	0.532 &	0.396 &	0.570 &	0.631 & pbr+real \\    \hline
\end{tabular}
\label{tab:bop challenge vergleich}
\vspace{-.2em}
\end{table}
\subsection{Ablation Studies}
Several systematic ablation studies have been conducted in this work to evaluate the influences of different components and parameters of the proposed method.
\vspace{-.7em}
\subsubsection*{PAFs:}
\begin{figure}[t]
 \centering
 \begin{subfigure}[t]{0.3\textwidth}
  \stackunder[-.2em]{\includegraphics[width=\textwidth]{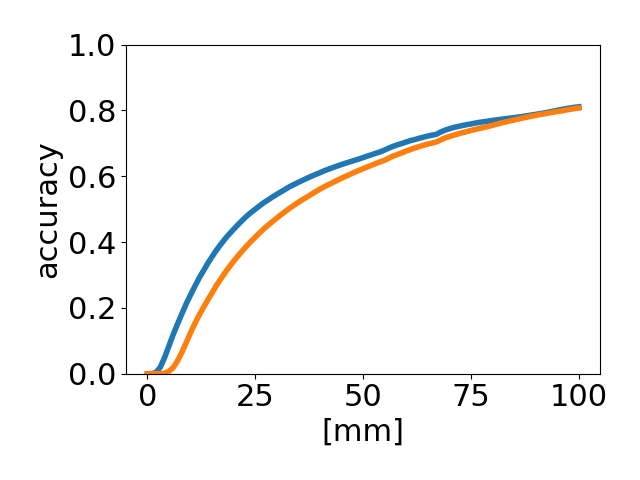}}{(a) ADD}
  \label{fig:paf_vs_heat_auc_add}
 \end{subfigure}
 \begin{subfigure}[t]{0.3\textwidth}
  \stackunder[-.2em]{\includegraphics[width=\textwidth]{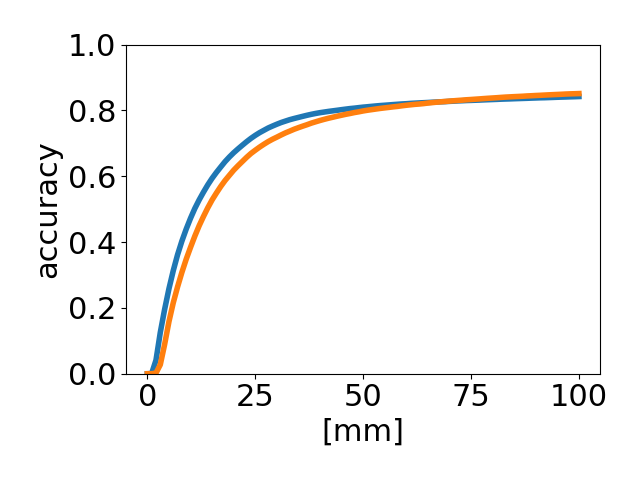}}{(b) ADD-S}
  \label{fig:paf_vs_heat_auc_adi}
 \end{subfigure}
  \begin{subfigure}[t]{0.3\textwidth}
  \stackunder[-.2em]{\includegraphics[width=\textwidth]{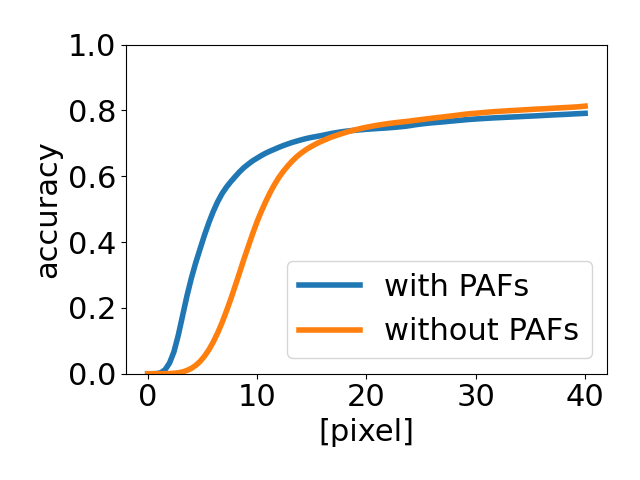}}{(c) 2D projection}
  \label{fig:paf_vs_heat_auc_rep}
 \end{subfigure}
 \vspace{-1em}
 \caption{Accuracy curves for all YCB-V objects using PAFs vs. heatmaps only.}
  \label{fig:paf_vs_heat_auc}
  \vspace{-1.5em}
\end{figure}

We investigate the benefit of PAFs for the YCB-V dataset, where a maximum of one instance per object class is present in an image.
For this, we infer the keypoints of an object instance as the global maximum of the respective heatmaps and do not use the PAFs computed by the CNN to assemble keypoints into object instances. This heatmaps-only approach is compared to the full approach using the PAF output. The results are presented in \reftab{tab:paf vs heat}. Using the PAFs improves the pose estimation result for almost all objects as well as the average accuracy. Without PAFs, recovering from wrong or ambiguous heatmap maxima is not possible, leading to inaccurate results especially in the case of occlusions and truncation.
An exception is Object 18, where the pose estimation is more accurate without using PAFs. This is probably due to the small size of the object leading to very short PAF vectors at the ends of the marker (cf. \reffig{fig:heatmaps_YCB-V Objekte}). These cannot be well detected by the model, leading to problems parsing the object skeleton.
\reffig{fig:paf_vs_heat_auc} shows the accuracy-threshold curves for each metric with and without using PAFs. The improvement using PAFs is most significant for small accuracy thresholds, demanding a precise estimation of the object pose. The two curves approach each other for higher thresholds.
\begin{table}[t]
\centering
\caption{AuC when using PAFs to connect keypoints into instances vs. only using global heatmap maxima. $^*$~denotes symmetric objects. Best results marked bold.}
\setlength{\tabcolsep}{7pt}
\begin{tabular}{|r||r|r|r||r|r|r|}
\hline
Object & \multicolumn{3}{c||}{using PAFs} & \multicolumn{3}{c|}{heatmaps only} \\
\hline
&ADD&ADD-S&2D Proj.&ADD&ADD-S&2D Proj.\\
\hline
 1 & \textbf{ 49.9} & \textbf{ 80.7} & \textbf{ 54.8} & 48.0 & 80.1 & 50.2 \\ 
 \hline 
 2 & \textbf{ 80.5} & \textbf{ 88.4} & \textbf{ 84.3} & 73.1 & 83.3 & 72.7 \\ 
 \hline 
 3 & \textbf{ 85.5} & \textbf{ 92.4} & \textbf{ 88.8} & 79.2 & 89.4 & 76.6 \\ 
 \hline 
 4 & \textbf{ 68.5} & \textbf{ 81.4} & \textbf{ 84.8} & 59.5 & 75.6 & 75.0 \\ 
 \hline 
 5 & \textbf{ 87.0} & \textbf{ 93.3} & \textbf{ 89.8} & 80.4 & 90.8 & 79.3 \\ 
 \hline 
 6 & \textbf{ 79.3} & \textbf{ 89.7} & \textbf{ 81.7} & 68.3 & 84.5 & 72.4 \\ 
 \hline 
 7 & \textbf{ 81.8} & \textbf{ 89.5} & \textbf{ 88.7} & 74.0 & 84.3 & 81.2 \\ 
 \hline 
 8 & \textbf{ 89.4} & \textbf{ 94.0} & \textbf{ 92.9} & 81.8 & 90.3 & 81.0 \\ 
 \hline 
 9 & \textbf{ 59.6} & \textbf{ 70.0} & \textbf{ 69.0} & 54.3 & 66.3 & 60.4 \\ 
 \hline 
10 & \textbf{ 36.5} & \textbf{ 58.3} & \textbf{ 55.0} & 35.2 & 55.3 & 53.0 \\ 
 \hline 
11 & \textbf{ 78.1} & \textbf{ 86.9} & \textbf{ 78.0} & 72.9 & 84.3 & 67.4 \\ 
 \hline 
12 & \textbf{ 56.7} & \textbf{ 67.1} & \textbf{ 66.2} & 51.7 & 64.3 & 57.8 \\ 
 \hline 
$^*$13 & \textbf{ 12.2} & \textbf{ 23.5} & \textbf{4.1} & 11.5 & 23.2& \textbf{  4.1}\\ 
 \hline 
14 & \textbf{ 54.0} & \textbf{ 76.9} & \textbf{ 75.2} & 49.1 & 72.0 & 63.3 \\ 
 \hline 
15 & \textbf{ 82.8} & \textbf{ 91.0} & \textbf{ 88.2} & 76.8 & 88.3 & 77.1 \\ 
 \hline 
$^*$16 & \textbf{ 16.7} & \textbf{ 29.6} & \textbf{ 29.5} & 15.7 & 27.4 & 23.2 \\ 
 \hline 
17 & \textbf{ 46.0} & \textbf{ 64.1} & \textbf{ 76.7} & 37.3 & 56.2 & 65.1 \\ 
 \hline 
18 & 9.8 & 11.9 & 20.8 & \textbf{ 36.1} & \textbf{ 43.1} & \textbf{ 65.1} \\ 
 \hline 
$^*$19 & \textbf{ 20.0} & \textbf{ 47.4} & 8.9 & 17.8 & 45.1 & \textbf{ 10.1} \\
 \hline
$^*$20 & \textbf{ 14.1} & \textbf{ 45.5} & 3.5 & 11.8 & 43.0 & \textbf{  4.7} \\
 \hline
$^*$21 & \textbf{ 12.1} & \textbf{ 29.7} & \textbf{  2.3} & 6.2 & 18.3 & 0.2\\
 \hline 
\hline
average& \textbf{59.0} & \textbf{72.7} & \textbf{65.0} & 54.5& 70.4& 58.9\\
\hline
\end{tabular}
\label{tab:paf vs heat}
\vspace{-1.3em}
\end{table}
\begin{table}[t]
\centering
\caption{AuC for manually and automatically picked keypoints.}
\setlength{\tabcolsep}{7pt}
\begin{tabular}{|r||c|c|c||c|c|c|}
\hline
Object&\multicolumn{3}{c||}{manually picked keypoints} & \multicolumn{3}{c|}{automatically picked keypoints} \\
\hline
&ADD&ADD-S&2D Proj.&ADD&ADD-S&2D Proj.\\    \hline
1 & \textbf{49.9} & \textbf{80.7} & 54.8 & 47.7 & 78.5 & \textbf{56.7} \\ \hline
2 & 80.5 & 88.4 & 84.3 & \textbf{81.0} & \textbf{89.8} & \textbf{87.1}\\    \hline
3 & \textbf{85.5} & \textbf{92.4} & 88.8 & 82.4 & 90.8 & \textbf{89.6}\\    \hline
4 & \textbf{68.5} & \textbf{81.4} & \textbf{84.8} & 68.4 & 80.8 & 84.3\\    \hline
5 & \textbf{87.0} & \textbf{93.3} & 89.8 & 83.7 & 92.0 & \textbf{92.0}\\    \hline
6 & \textbf{79.3} & \textbf{89.7} & 81.7 & 77.3 & 88.9 & \textbf{85.2}\\    \hline
7 & \textbf{81.8} & \textbf{89.5} & \textbf{88.7} & 75.9 & 85.0 & 85.2\\    \hline
8 & \textbf{89.4} & \textbf{94.0} & \textbf{92.9} & 81.0 & 89.6 & 92.4\\    \hline
\end{tabular}
\label{tab:auto vs handpicked keyp}
\vspace{-1.em}
\end{table}
\begin{table}[!ht]
\centering
\caption{AuC for ADD(-S) and 2D-projection metrics comparing models trained to detect one object class versus models detecting two object classes.}
\setlength{\tabcolsep}{7pt}
\begin{tabular}{|r||r|r|r||r|r|r|}
\hline
Object & \multicolumn{3}{c||}{1-Object models} & \multicolumn{3}{c|}{2-Object models} \\
\hline
&ADD&ADD-S&2D Proj.&ADD&ADD-S&2D Proj.\\
\hline
 1 & \textbf{ 49.9} & \textbf{ 80.7} & \textbf{ 54.8} & 18.0 & 31.5 & 23.0 \\    \hline 
 4 & \textbf{ 68.5} & \textbf{ 81.4} & \textbf{ 84.8} & 38.5 & 43.6 & 46.8 \\    \hline 
 5 & \textbf{ 87.0} & \textbf{ 93.3} & \textbf{ 89.8} & 20.3 & 23.8 & 23.9 \\    \hline 
 6 & \textbf{ 79.3} & \textbf{ 89.7} & \textbf{ 81.7} & 53.6 & 63.1 & 60.3 \\    \hline 
 7 & \textbf{ 81.8} & \textbf{ 89.5} & \textbf{ 88.7} & 38.5 & 42.3 & 42.5 \\    \hline 
 8 & \textbf{ 89.4} & \textbf{ 94.0} & \textbf{ 92.9} & 39.8 & 44.4 & 45.3 \\    \hline 
 9 & \textbf{ 59.6} & \textbf{ 70.0} & \textbf{ 69.0} & 25.2 & 33.3 & 35.5 \\    \hline 
10 & \textbf{ 36.5} & \textbf{ 58.3} & \textbf{ 55.0} & 8.3 & 12.4 & 16.9 \\    \hline 
\hline
average&\textbf{66.2}&\textbf{81.3}&\textbf{75.2}&32.2&39.8&39.2\\
\hline
\end{tabular}
\label{tab:1 obj models vs 2 obj models}
\vspace{-.5em}
\end{table}
\subsubsection*{Selection of Keypoints on Object Models:}
In \reftab{tab:auto vs handpicked keyp}, we compare evaluation results using manually and automatically chosen object keypoints for an exemplary subset of the object classes. The estimated pose generally is more accurate using the manually defined keypoints. As discussed in \refsec{sec:Method}, these keypoints are easier to find and can be more precisely localized by the CNN architecture, as they are placed on distinct spots of the object geometry and texture (cf. Figs.~\ref{fig:heatmaps_YCB-V Objekte} and~\ref{fig:heatmaps_autoYCB-V Objekte}). The locations of the automatically chosen keypoints, on the other hand, are often weakly constrained along edges or on the object surface, thus being predicted less precisely.
\begin{table}[t]
\centering
\caption{AuC using a 2-object model with and without rescaling layer width.}
\setlength{\tabcolsep}{7pt}
\begin{tabular}{|r||r|r|r||r|r|r|}
\hline
Object & \multicolumn{3}{c||}{doubled layer width} & \multicolumn{3}{c|}{normal layer width} \\
\hline
&ADD&ADD-S&2D Proj.&ADD&ADD-S&2D Proj.\\
\hline
 1 & \textbf{ 19.1} & \textbf{ 31.5} & \textbf{ 25.8} & 18.0 & \textbf{31.5} & 23.0 \\    \hline 
 4 & \textbf{ 41.5} & \textbf{ 47.1} & \textbf{ 47.4} & 38.5 & 43.6 & 46.8 \\    \hline 
\hline
average&\textbf{32.2}&\textbf{40.7}&38.5& \textbf{32.2} &39.8&\textbf{39.2}\\
\hline
\end{tabular}
\label{tab:1 obj models vs 2 obj models size 2}
\vspace{-.5em}
\end{table}
\begin{figure}
 \centering
 \begin{subfigure}[t]{0.3\textwidth}
  \stackunder[-.2em]{\includegraphics[width=\textwidth]{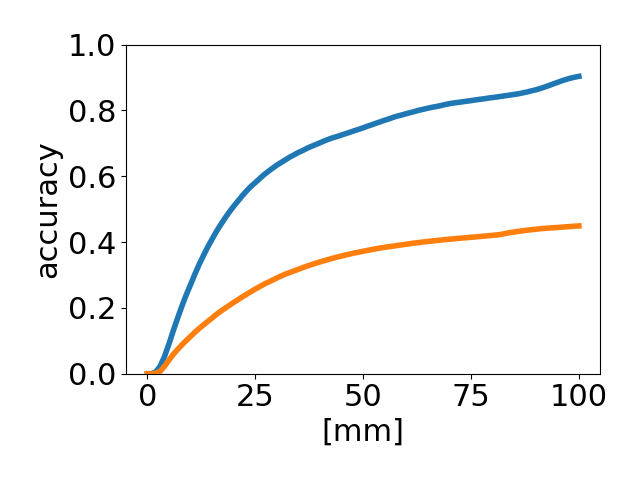}}{(a) ADD}
  \label{fig:1_obj_vs_2_obj_auc_add}
 \end{subfigure}
 \begin{subfigure}[t]{0.3\textwidth}
  \stackunder[-.2em]{\includegraphics[width=\textwidth]{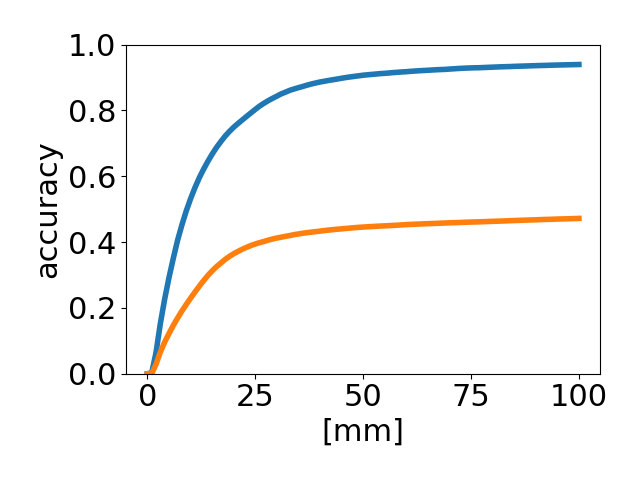}}{(b) ADD-S}
  \label{fig:1_obj_vs_2_obj_auc_adi}
 \end{subfigure}
  \begin{subfigure}[t]{0.3\textwidth}
  \stackunder[-.2em]{\includegraphics[width=\textwidth]{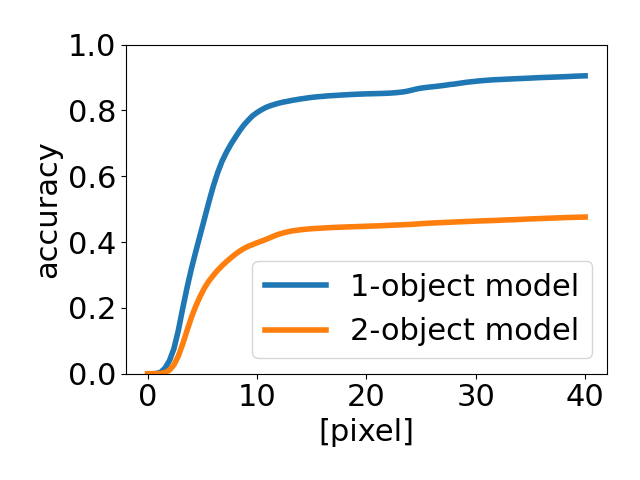}}{(c) 2D projection}
  \label{fig:1_obj_vs_2_obj_auc_rep}
 \end{subfigure}
 \vspace{-1em}
 \caption{Accuracy curves for all YCB-V objects using models that detect one object class vs. models detecting two object classes.}
  \label{fig:1_obj_vs_2_obj_auc}
  \vspace{-1em}
\end{figure}
\vspace{-.3em}
\subsubsection*{Number of Object Classes per Model:}
We also investigate the influence of training the CNN model to detect keypoints and PAFs for objects of one or of several classes. The results of this comparison are shown in \reftab{tab:1 obj models vs 2 obj models} exemplary for a subset of the object classes. The performance of models trained for one object class only is significantly better. The plot of the accuracy-threshold curves shown in \reffig{fig:1_obj_vs_2_obj_auc} confirms these results. Also, rescaling the width of the CNN model, i.e., doubling the number of channels at each stage of the network, does not significantly improve the results of the 2-object model, as is shown in \reftab{tab:1 obj models vs 2 obj models size 2}.
Therefore, a separate model per object class has been used in this work.
 
\section{Conclusion}
\label{sec:Conclusion}
In this work, we introduce a pipeline for 6D object pose estimation adopting the OpenPose CNN architecture~\cite{OpenPose} to predict keypoints and Part Affinity Fields (PAFs) on objects. Keypoints are defined on prominent geometric features of the object contour (i.e., corners) and interconnected by PAFs to form a cuboid-like structure. They are predicted as local maxima of a pixel-wise heatmap and assembled into instances using the PAFs. This bottom-up approach permits to infer keypoints directly from the input image, without prior object detection or segmentation. Object poses are then calculated via the PnP-RANSAC algorithm using 2D-3D correspondences between detected and model keypoints.
The proposed approach is evaluated on the YCB-Video dataset, containing 21 typical household objects important for domestic service robot applications. Our method achieves accuracy comparable to recent state-of-the-art methods using RGB images only, without employing further pose refinement. The usage of PAFs is shown to be advantageous compared to directly using global maxima of the heatmaps as keypoint detections. Models trained for a single object class perform significantly better than models trained for multiple object classes.

Directions for future work include improved handling of symmetric objects and evaluating the method in scenarios where multiple instances of the same object class occur. Furthermore, a method for automatic keypoint and PAF selection without sacrificing accuracy compared to manual selection should be investigated, enabling to efficiently extend the method with novel object models. 
\section*{Acknowledgments}
This work was funded by grant BE 2556/18-2 of the German Research Foundation (DFG).

\bibliographystyle{splncs04}
\bibliography{paper}

\end{document}